\documentclass[conference]{IEEEtran}

\usepackage[utf8]{inputenc}
\usepackage{cite}
\usepackage{amsmath,amssymb,amsfonts}
\usepackage{amsthm}
\usepackage{algorithm}
\usepackage{algorithmic}
\usepackage{array}
\usepackage{booktabs}
\usepackage{graphicx}
\usepackage{listings}
\usepackage{multirow}
\usepackage{newfloat}
\usepackage{tabularx}
\usepackage{tcolorbox}
\usepackage{textcomp}
\usepackage{xcolor}
\usepackage{url}
\usepackage[hidelinks]{hyperref}
\tcbuselibrary{listings, breakable}

\def\BibTeX{{\rm B\kern-.05em{\sc i\kern-.025em b}\kern-.08em
    T\kern-.1667em\lower.7ex\hbox{E}\kern-.125emX}}

\lstdefinelanguage{SQL}{
  morekeywords={
    SELECT, FROM, WHERE, AND, OR, NOT, INSERT, INTO, VALUES, UPDATE, SET, DELETE,
    CREATE, TABLE, DROP, ALTER, ADD, JOIN, ON, AS, DISTINCT, COUNT, RETURN, MATCH,
    OPTIONAL, WITH, UNION, ALL, GROUP, BY, ORDER, LIMIT, OFFSET
  },
  sensitive=false,
  morecomment=[l]{--},
  morestring=[b]',
}

\tcbset{
  colback=gray!5,
  colframe=black,
  boxsep=1pt,
  left=3pt,
  right=3pt,
  top=3pt,
  bottom=3pt,
  before skip=4pt,
  after skip=4pt,
}

\floatstyle{ruled}
\newfloat{listing}{tb}{lst}{}
\floatname{listing}{Listing}

\title{VeriCypher: Named-Entity Verification for GraphRAG via Text-to-Cypher Generation}

\author{%
\IEEEauthorblockN{Anton Gusarov\IEEEauthorrefmark{1}, Anastasia Volkova\IEEEauthorrefmark{1}, Valentin Khrulkov\IEEEauthorrefmark{1}\IEEEauthorrefmark{2},\\
Evgenii Maslov\IEEEauthorrefmark{1}, Andrey Kuznetsov\IEEEauthorrefmark{1}\IEEEauthorrefmark{3}, and Ivan Oseledets\IEEEauthorrefmark{4}}
\IEEEauthorblockA{%
\IEEEauthorrefmark{1}\textit{FusionBrain Lab, Artificial Intelligence Research Institute}\\
\IEEEauthorrefmark{2}\textit{Research Center of the Artificial Intelligence Institute, Innopolis University}\\
\IEEEauthorrefmark{3}\textit{Multimodal Industrial GenAI Lab, Innopolis University}\\
\IEEEauthorrefmark{4}\textit{Artificial Intelligence Research Institute}}
\IEEEauthorblockA{Corresponding author: Anton Gusarov}}

\begin{document}
\bstctlcite{IEEEexample:BSTcontrol}

\maketitle

\begin{abstract}
Knowledge graph question answering has traditionally focused on Resource Description Framework (RDF) graphs, SPARQL queries, and triple-based representations. Labeled Property Graphs (LPGs), however, pose a different challenge for LLM-based query generation: executable Cypher queries must align with schema structure, traversal direction, property keys, and literal entity values. This motivated us to introduce \textit{VeriCypher}, an LLM-based text-to-Cypher framework for question answering over LPGs. It combines query generation conditioned on Cypher-like graph schema summaries with runtime execution, LLM-based query evaluation, and iterative correction where the central mechanism is named-entity verification: the system extracts node labels, relationship patterns, property keys, and literal values from generated queries, checks them against the target graph and schema, and turns mismatches into correction instructions for the next generation step. We evaluate \textit{VeriCypher} on the CypherBench benchmark across multiple domains and LLM backbones. Compared with an LLM baseline, this method consistently improves answer correctness and query robustness. Ablation results further show that named-entity verification helps correct hallucinated schema elements and entity values, supporting more reliable LLM-based question answering over heterogeneous property graphs. The \textit{VeriCypher} implementation and experimental results are available at \url{https://anonymous.4open.science/r/veri-cypher/}.
\end{abstract}

\begin{IEEEkeywords}
text-to-Cypher, graph question answering, property graphs, large language models, named-entity verification
\end{IEEEkeywords}

\section{Introduction}

Graph databases provide a model for representing richly connected data through labeled nodes, typed relationships, and key--value properties \cite{Angles2018ThePG}. They are increasingly adopted across various domains such as biomedicine \cite{Walsh2020BioKGAK} and infrastructure management \cite{Donkers2020LinkedDF}, where information access often requires reasoning over heterogeneous entities and multi-hop relations. In this work, we study how large language models (LLMs) can support natural-language question answering over such graph-structured data by generating executable Cypher queries.

Most prior work on knowledge graph question answering (KGQA) targets Resource Description Framework (RDF) graphs, where data are represented as triples and queried with SPARQL. Labeled Property Graphs (LPGs), by contrast, use labeled nodes, typed relationships, and first-class properties on both nodes and edges. This modeling choice makes LPGs well suited for heterogeneous, attribute-rich graph data, but also changes the query-generation problem: systems must synthesize traversal-oriented Cypher queries rather than triple-pattern SPARQL queries.

These differences make natural-language access to LPGs challenging. A Cypher query must combine the correct node labels, relationship types, property keys, literal values, and traversal direction, while also matching the intent of the user question. Errors in any of these elements can lead to syntactically valid but incorrect queries, empty results, or runtime failures. Although text-to-query generation has been extensively studied for SQL and RDF/SPARQL, LLM-based synthesis of executable Cypher over heterogeneous property graphs remains comparatively underexplored.

We address this gap by introducing \textit{VeriCypher}, an LLM-based text-to-Cypher framework that combines execution feedback, query evaluation, and named-entity verification within an iterative correction loop. \textit{VeriCypher} checks generated node labels, relationship patterns, property keys, and literal values against the target property graph and schema, then converts detected mismatches into correction instructions for subsequent query generation. We evaluate the approach on CypherBench \cite{DBLP:conf/acl/FengPR25} across multiple domains and LLM backbones, comparing the full workflow against a linear LLM baseline used in CypherBench \cite{DBLP:conf/acl/FengPR25} and an ablation without named-entity verification.

\subsection{Labeled Property Graphs}

We model a labeled property graph (LPG) as a directed multigraph with labels, relationship types, and key--value properties \cite{angles2024}. Formally, a property graph is a tuple
\[
\mathcal{G} = (N, E, \rho, \lambda_N, \lambda_E, \sigma)
\]
where:
\begin{itemize}
  \item[--] $N$ is a finite set of nodes;
  \item[--] $E$ is a finite set of edges, disjoint from $N$;
  \item[--] $\rho: E \rightarrow N \times N$ is a total function assigning each edge its source and target nodes;
  \item[--] $\lambda_N : N \rightarrow \mathcal{P}_{\mathrm{fin}}^{+}(L_N)$ assigns each node a finite non-empty set of labels from a set of node labels $L_N$;
  \item[--] $\lambda_E : E \rightarrow L_E$ assigns each edge a relationship type from a set of relationship types $L_E$;
  \item[--] $\sigma : (N \cup E) \times P \rightharpoonup V$ is a partial function assigning property values from $V$ to node or edge property keys from $P$.
\end{itemize}

A directed path instance, allowing repeated nodes and edges as in Cypher path values, is a finite alternating sequence
\[
\pi = (n_1, e_1, n_2, e_2, \dots, e_k, n_{k+1})
\]
such that $k \geq 0$, $n_i \in N$, $e_i \in E$, and $\rho(e_i) = (n_i, n_{i+1})$ for all $1 \leq i \leq k$. Its length is $k$, i.e. the number of traversed edges. For $k=0$, the walk consists of a single node.

The relationship-type sequence of $\pi$ is
\[
\mathrm{type}(\pi) = (\lambda_E(e_1), \ldots, \lambda_E(e_k)).
\]
For $k = 0$, this sequence is empty.

% ================================================================================================
\section{Related Work}
% ================================================================================================

\begin{figure*}[ht]
\centering
\includegraphics[width=0.9\textwidth]{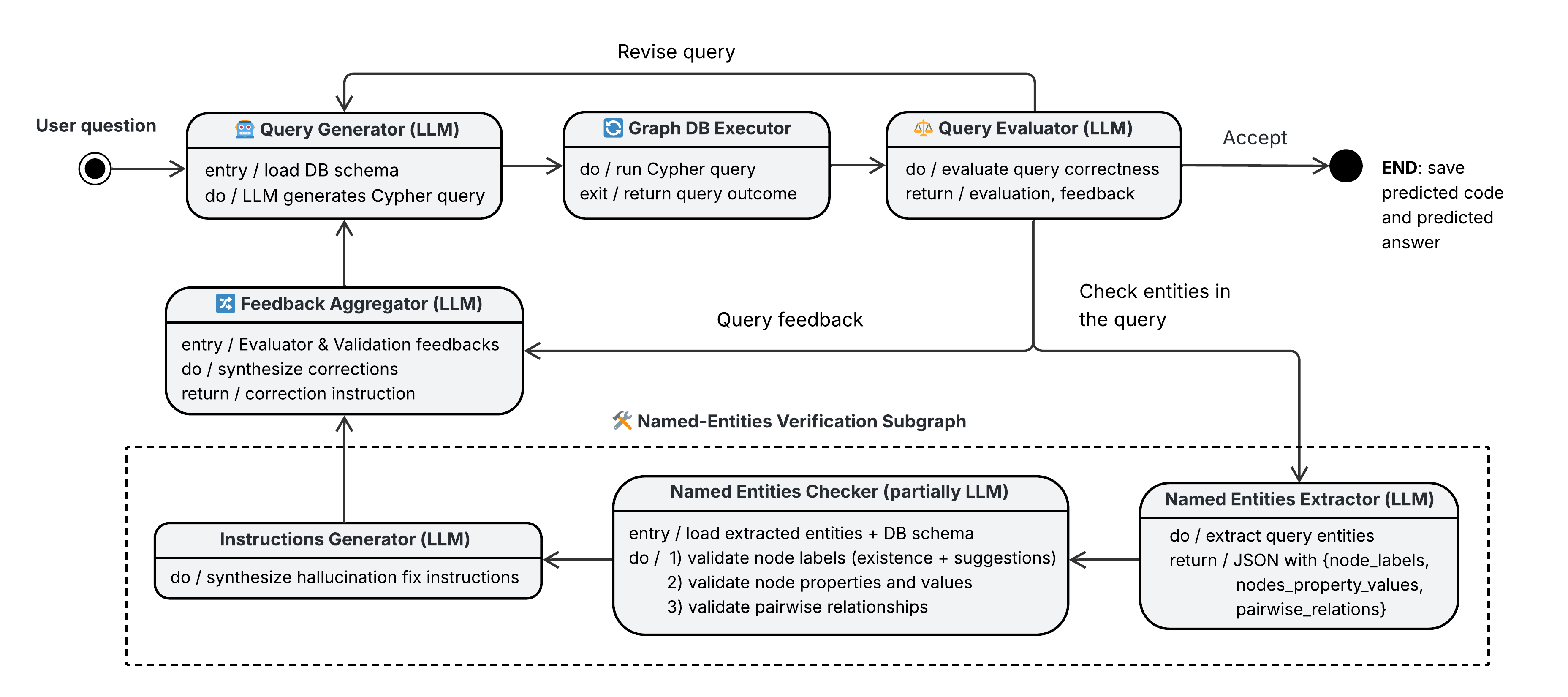}
\caption{State diagram of the \textit{VeriCypher} workflow for text-to-Cypher generation over labeled property graphs (LPGs), including query generation, execution, evaluation, named-entity verification, and feedback-driven refinement.}
\label{architecture}
\end{figure*}

%------------------------------------

Cypher is a widely adopted declarative query language for property graphs, originally developed for Neo4j and later opened through the openCypher project \cite{francis2018cypher}. Its pattern-matching syntax is closely related to recent property-graph query language standardization, including ISO/IEC 39075:2024 GQL \cite{iso39075}. Nevertheless, LLM-based question answering over property graphs remains less studied than RDF/SPARQL-based knowledge graph question answering. Existing property-graph research has largely focused on query language design, algebraic foundations, interoperability, and query evaluation, while fewer studies examine executable Cypher synthesis under schema, entity, and runtime constraints.

\subsection{Knowledge Graph Question Answering}

Knowledge graph question answering (KGQA) aims to answer natural-language questions over structured graph-based knowledge sources. Prior work commonly distinguishes semantic parsing approaches, which translate questions into executable logical forms or query languages, from information-retrieval approaches, which retrieve relevant entities, relations, paths, or subgraphs before answer generation \cite{DBLP:conf/ijcai/LanHJ0ZW21,DBLP:conf/acl/LuoETPG0MDSLZL24,DBLP:conf/iclr/JiangZ0W23,DBLP:conf/acl/JiangZZS0ZW25}. Recent studies also evaluate LLMs as semantic parsers for graph-query generation, showing both their promise and their sensitivity to schema and query-form constraints \cite{DBLP:conf/icaart/SchneiderKJSM24}. Most of this literature targets RDF/SPARQL-style KGQA, whereas our work focuses on executable Cypher generation over labeled property graphs (LPG).

\subsection{Text-to-SQL Semantic Parsing}

Natural language interfaces to relational databases are a long-standing line of semantic parsing research and provide important methodological context for the text-to-Cypher class of problems. Early neural systems such as Seq2SQL \cite{DBLP:journals/corr/abs-1709-00103} and SQLNet \cite{DBLP:journals/corr/abs-1711-04436} targeted WikiSQL-style queries, combining sequence generation, sketch prediction, and slot filling to reduce the output search space. TypeSQL \cite{DBLP:conf/naacl/YuLZZR18} further incorporated type information and external knowledge to improve schema and value linking.

The introduction of Spider \cite{DBLP:conf/emnlp/YuZYYWLMLYRZR18} shifted the field toward cross-domain generalization, where models must map questions to previously unseen database schemas. Subsequent structure-aware parsers explicitly model this schema dependence. SyntaxSQLNet \cite{DBLP:journals/corr/abs-1810-05237} uses a syntax-tree decoder guided by SQL grammar, while RAT-SQL \cite{DBLP:conf/acl/WangSLPR20} represents the question and schema as a relation-aware graph to improve schema linking. SmBoP \cite{DBLP:conf/acl-spnlp/RubinB21} then generates SQL through a semi-autoregressive bottom-up decoder over compositional query trees. More recent systems such as RESDSQL \cite{DBLP:conf/aaai/Li00023} decouple schema linking from SQL skeleton parsing, reinforcing the view that grounding query tokens in schema elements is a central source of difficulty.

Another relevant thread constrains or verifies generated programs after decoding. PICARD \cite{scholak2021picard} incrementally rejects inadmissible tokens during autoregressive generation, improving syntactic validity for SQL. LLM-based approaches also use decomposition, execution feedback, and database-content grounding; DIN-SQL \cite{DBLP:conf/nips/PourrezaR23} decomposes text-to-SQL prompting with self-correction, while BIRD \cite{DBLP:conf/nips/LiHQYLLWQGHZ0LC23} highlights that real-world text-to-SQL requires grounding not only in schema but also in database values. 

These findings motivate our treatment of text-to-Cypher as an execution-grounded generation problem. Unlike SQL parsers, however, \textit{VeriCypher} must verify graph-specific entities: node labels, relationship types and directions, property keys, and literal values in a labeled property graph.

\subsection{Agentic and Search-Based LLMs for Text-to-SQL} 
% ------------------------------------------
Recent LLM-based Text-to-SQL systems increasingly move beyond single-pass prompting toward test-time decomposition, candidate search, and refinement. CHASE-SQL \cite{DBLP:conf/iclr/PourrezaL0CTKGS25} generates diverse SQL candidates through complementary reasoning paths, including divide-and-conquer decomposition, execution-plan reasoning, and instance-aware demonstrations, followed by learned pairwise candidate selection. MAC-SQL \cite{DBLP:conf/coling/WangR0LBCYZYSL25} uses a decomposer agent supported by auxiliary modules for sub-database selection and SQL refinement, addressing large schemas and complex questions. Alpha-SQL \cite{DBLP:conf/icml/Li0FXC0L25} formulates zero-shot Text-to-SQL as a search problem, applying Monte Carlo Tree Search over partial SQL states with an LLM-based action model and a self-supervised reward function.

\textit{VeriCypher} adopts the same broader insight of execution-aware iterative refinement, but applies it to Cypher over labeled property graphs, where correction must also verify graph-specific entities such as node labels, relationship types and directions, property keys, and literal values.

\subsection{Agent-Based Graph Reasoning}
% ------------------------------------------
Graph-based retrieval-augmented generation and graph reasoning methods use explicit graph structure as an external source of evidence for LLMs \cite{DBLP:journals/corr/abs-2501-00309,DBLP:journals/tois/PengZLBSHZT26}. Graph Chain-of-Thought (Graph-CoT) \cite{DBLP:conf/acl/JinXZRZL0TWM024} proposes iterative reasoning over graph data, alternating between LLM reasoning, graph interaction, and graph execution. Graph Counselor \cite{DBLP:conf/acl/GaoZALNQL25} extends this direction with multiple cooperating agents for adaptive graph information extraction and multi-perspective self-reflection.

These methods demonstrate the value of iterative interaction with graph-structured evidence, but they differ from our objective. Most graph-reasoning and GraphRAG systems retrieve, traverse, or summarize graph context to support answer generation. \textit{VeriCypher} instead treats the generated Cypher query as the object to be executed, diagnosed, and repaired. The graph database is therefore not only a retrieval substrate but also a verification environment used to detect hallucinated schema elements, invalid relationship patterns, and incorrect property values before the final answer is produced.

\subsection{Text-to-Cypher}
% ------------------------------------------
Text-to-Cypher is a comparatively recent specialization of natural-language interfaces for property-graph databases. Hornsteiner et al.~\cite{DBLP:journals/fi/HornsteinerKSEES24} develop a modular natural language interface for Neo4j, combining Cypher generation, database selection, and error correction in a few-shot setting. Chatterjee and Dethlefs~\cite{windTurbine} study a domain-specific QA system for wind-turbine maintenance, where a semantic parser maps natural-language questions to Cypher for decision-making support.

Dataset and benchmark work has also started to address the shortage of reusable Text-to-Cypher resources. SpCQL \cite{DBLP:conf/cikm/GuoLXT022} provides 10,000 question--Cypher pairs over an OwnThink-based Neo4j graph, while CypherBench \cite{DBLP:conf/acl/FengPR25} evaluates LLM-based question answering over full-schema Wikidata-derived property graphs. Synthetic-data pipelines such as SyntheT2C \cite{DBLP:conf/coling/ZhongZSJQZ25} and Auto-Cypher \cite{DBLP:conf/naacl/TiwariMYHM25} further address data scarcity by generating question--Cypher pairs for fine-tuning and evaluation.

Recent inference-time methods begin to use execution feedback for Text-to-Cypher. Reflection-augmented scaling (RAS) \cite{DBLP:journals/corr/abs-2605-22937} shows that conditioning repeated Cypher generation attempts on prior database error messages can improve executability compared with independent resampling. This direction is close to our use of runtime feedback, but focuses primarily on reducing execution errors rather than verifying whether graph-specific names and values are grounded in the target database.

In comparison to prior work focused on prompt-based generation, Neo4j interfaces, benchmarks, synthetic data, or execution-error reduction, \textit{VeriCypher} treats Text-to-Cypher as an execution-grounded iterative-refinement problem, combining named-entity verification, a Memgraph backend, and semantic--syntactic feedback to repair graph-specific failures such as hallucinated node labels, relationship types and directions, property keys, and literal values.

% ================================================================================================
\section{Implementation}\label{sec:implementation}
% ================================================================================================

We implement \textit{VeriCypher} as a modular LLM workflow for text-to-Cypher over LPGs. The workflow augments single-pass query generation with execution feedback, LLM-based query evaluation, and named-entity verification, allowing generated queries to be iteratively revised when they contain syntax or runtime errors, empty-result failures, hallucinated schema elements, or mismatched entity values. This feedback-driven refinement follows the broader observation that LLMs can improve generated programs through iterative correction \cite{DBLP:conf/nips/QuZGK24}, but grounds the correction process in graph-specific evidence from the target database and schema. Figure~\ref{architecture} provides an overview of the system components and data flow. The full implementation, including experimental outputs and metric artifacts, is available at: \url{https://anonymous.4open.science/r/veri-cypher/}.

\paragraph{Workflow Components} The full \textit{VeriCypher} workflow is composed of the following nodes, each specializing in a distinct subtask.

\begin{enumerate}
    \item \textbf{\textit{Query Generator}} formulates a Cypher query based on the user question in natural language. The generation is grounded in the graph schema (in a form close to Cypher syntax), encompassing both node specifications and pairwise relationship descriptions. In subsequent iterations, if the first-shot generation was not accepted, the generator uses evaluator feedback or aggregated validation feedback to refine the Cypher query generated in the previous iteration.
    
    \item \textbf{\textit{Graph Database Executor}} interfaces with the underlying graph database engine (Memgraph in our implementation) to execute the generated Cypher query and retrieve the query outcome, which may consist of structured result data, an error message, or an empty result. Empty results commonly arise from wrong entity values, invalid traversal paths, overly strict filters, or schema-name hallucinations.
    
    \item \textbf{\textit{Query Evaluator}} is responsible for assessing the semantic and logical adequacy of the generated Cypher query relative to the user’s intent. 
    Functionally, it serves as an \textit{LLM-based critic} \cite{DBLP:journals/corr/abs-2407-00215,DBLP:conf/nips/YangYLYZTLY25} within our pipeline. The evaluator is prompted to analyze three key aspects: (1) consistency between the user's intended semantics and the explanation of the generated query, (2) alignment of the query logic with the user question, and (3) validity and informativeness of the returned results. It returns structured feedback and one of three query grades:
    \begin{itemize}
        \item \textbf{Accept}: the query is error-free and the returned results fully and logically answer the user question.
        \item \textbf{Incorrect}: the query executes without error and returns data, but is semantically misaligned, logically flawed, or incomplete.
        \item \textbf{Error or empty}: the query either fails to execute due to a runtime error or returns no results.
    \end{itemize}
    The grade is then mapped to a workflow route. \textbf{Accept} terminates the graph, \textbf{Incorrect} routes to direct query revision, and \textbf{Error or empty} routes either to direct revision or to \textbf{Check named entities}, depending on whether the failure appears to be a Cypher-construction error or a graph-name/entity mismatch.
    
    \item \textbf{\textit{Named Entity Extractor}} is an LLM-backed component that identifies elements within the query such as node labels, property-value pairs, and pairwise relationship patterns that are susceptible to hallucination. Its primary function is to decompose the query so these elements can subsequently be verified against the underlying graph data.
    
    \item \textbf{\textit{Verification Module}} follows the Named Entity Extractor and verifies the extracted labels, property values, and pairwise relationship patterns against the graph and schema files. This is done programmatically via auxiliary Cypher queries and schema lookup. For missing or suspicious entities, the module retrieves candidate replacements using normalized Levenshtein similarity \cite{rapidfuzz} and, when applicable, uses an LLM to select semantically relevant alternatives.
    
    \item \textbf{\textit{Instructions Generator}} synthesizes revision instructions based on the verification results targeting hallucinated or misnamed entities. It takes as input the structured feedback from the Verification Module, including which entities failed to match the data content and which alternatives were suggested. The output is a concise instruction to guide the Query Generator on how to revise the query.
    
    \item \textbf{\textit{Feedback Aggregator}} integrates the outputs of both the Query Evaluator and the Verification Module into a unified correction strategy. It consolidates signals such as semantic inconsistencies, execution errors, and naming mismatches to produce structured and prioritized feedback. This aggregated feedback is the basis for guiding the subsequent correction of the query, ensuring that both logical soundness and schema compliance are addressed.
    
    \item Accepted queries terminate the graph and are saved together with their executed Memgraph results as \texttt{predicted\_code} and \texttt{predicted\_answer}.
\end{enumerate}

Each LLM-backed component is guided by a system prompt tailored to its role. The \textbf{\textit{Query Generator}} uses the same prompt as in \cite{DBLP:conf/acl/FengPR25}. The complete prompt templates are given in the public implementation: \url{https://anonymous.4open.science/r/veri-cypher/}.

\begin{algorithm}[t]
\caption{\textit{VeriCypher} query refinement with execution feedback and named-entity validation}
\label{alg:cypher_refinement}
\footnotesize
\textbf{Input}: question $Q$, graph database $G$, schema $S$, attempt budget $B$ \\
\textbf{Output}: selected Cypher query $C$ and executed result $R$ \\
\begin{algorithmic}[1]
\STATE $C \leftarrow \mathrm{GenerateQuery}(Q, S)$
\STATE $H \leftarrow \emptyset$ \quad \textit{// attempts history}
\FOR{$t \leftarrow 1$ \TO $B$}
    \STATE $R \leftarrow \mathrm{ExecuteQuery}(C)$
    \STATE $(E, r) \leftarrow \mathrm{EvaluateAndRoute}(Q, C, R)$
    \STATE $H \leftarrow \mathrm{Append}(H, (C, R, E, r))$
    \IF{$r = \textsc{Accept}$}
        \STATE \textbf{return} $C$, $R$
    \ELSIF{$t = B$}
        \STATE \textbf{break}
    \ENDIF
    \IF{$r = \textsc{ReviseQuery}$}
        \STATE $C \leftarrow \mathrm{GenerateQuery}(Q, S, E_{feedback})$
    \ELSIF{$r = \textsc{ValidateEntities}$}
        \STATE $E_{ent} \leftarrow \mathrm{ExtractNamedEntities}(C)$
        \STATE $L \leftarrow \mathrm{CheckLabels}(E_{ent}, G)$
        \STATE $P \leftarrow \mathrm{CheckProps}(E_{ent}, G)$
        \STATE $M \leftarrow \mathrm{CheckPatterns}(E_{ent}, S)$
        \STATE $V \leftarrow \mathrm{LevenshteinSuggest}(L, P, M)$
        \STATE $V \leftarrow \mathrm{LLMPropSuggest}(V)$
        \STATE $I \leftarrow \mathrm{FixInstructions}(V)$
        \STATE $F \leftarrow \mathrm{AggregateFeedback}(Q, E_{feedback}, I)$
        \STATE $C \leftarrow \mathrm{GenerateQuery}(Q, S, F)$
    \ENDIF
\ENDFOR
\STATE \textbf{return} $\mathrm{SelectSavedResult}(H)$
\end{algorithmic}
\end{algorithm}

\begin{table*}[!t]
\caption{CypherBench performance by graph, model, and workflow; values are per-graph average percentages.}
\label{table:workflow_metric_comparison}
\centering
\footnotesize
\begin{tabular}{@{}llcccccc@{}}
\toprule
 &  & \multicolumn{3}{c}{\textbf{DeepSeek V4 Flash}} & \multicolumn{3}{c}{\textbf{Qwen3 Coder Next}} \\
\cmidrule(lr){3-5}\cmidrule(l){6-8}
\textbf{Graph} & \textbf{Metric} & \textbf{Single} & \shortstack{\textbf{VeriCypher}\\\textbf{(no entity check)}} & \textbf{VeriCypher} & \textbf{Single} & \shortstack{\textbf{VeriCypher}\\\textbf{(no entity check)}} & \textbf{VeriCypher} \\
\midrule
\multirow{4}{*}{company} & EX & 97.12\% & 98.56\% & 97.69\% & 90.78\% & 98.56\% & 94.59\% \\
 & PSJS & 65.23\% & 82.80\% & 82.74\% & 50.78\% & 63.55\% & 71.62\% \\
 & AEM & 59.37\% & 64.84\% & 69.45\% & 41.21\% & 51.01\% & 56.25\% \\
 & Set F1 & 73.46\% & 77.16\% & 79.23\% & 47.75\% & 59.62\% & 66.58\% \\
\midrule
\multirow{4}{*}{\shortstack{fictional\\character}} & EX & 96.62\% & 98.44\% & 98.70\% & 92.73\% & 92.99\% & 94.55\% \\
 & PSJS & 76.83\% & 83.51\% & 85.80\% & 61.07\% & 70.72\% & 71.86\% \\
 & AEM & 60.78\% & 66.75\% & 70.65\% & 42.86\% & 54.03\% & 54.81\% \\
 & Set F1 & 74.54\% & 80.05\% & 83.57\% & 56.84\% & 68.53\% & 69.77\% \\
\midrule
\multirow{4}{*}{\shortstack{flight\\accident}} & EX & 94.71\% & 98.94\% & 100.00\% & 94.71\% & 98.41\% & 99.47\% \\
 & PSJS & 87.10\% & 89.47\% & 91.21\% & 83.34\% & 89.42\% & 91.85\% \\
 & AEM & 75.13\% & 78.31\% & 80.95\% & 67.20\% & 74.60\% & 78.31\% \\
 & Set F1 & 82.09\% & 84.30\% & 88.97\% & 77.61\% & 84.59\% & 87.94\% \\
\midrule
\multirow{4}{*}{geography} & EX & 95.36\% & 99.45\% & 99.45\% & 90.98\% & 78.96\% & 96.45\% \\
 & PSJS & 83.16\% & 85.30\% & 86.24\% & 72.83\% & 75.56\% & 76.35\% \\
 & AEM & 62.30\% & 68.03\% & 72.13\% & 53.83\% & 56.28\% & 54.64\% \\
 & Set F1 & 79.09\% & 78.49\% & 82.80\% & 69.02\% & 69.07\% & 68.64\% \\
\midrule
\multirow{4}{*}{movie} & EX & 95.76\% & 99.50\% & 98.75\% & 86.53\% & 96.51\% & 96.26\% \\
 & PSJS & 84.34\% & 89.13\% & 90.18\% & 63.43\% & 72.06\% & 75.52\% \\
 & AEM & 69.33\% & 75.31\% & 76.81\% & 46.38\% & 55.61\% & 59.85\% \\
 & Set F1 & 81.88\% & 84.56\% & 86.22\% & 58.03\% & 67.34\% & 70.85\% \\
\midrule
\multirow{4}{*}{NBA} & EX & 97.04\% & 92.59\% & 98.52\% & 91.48\% & 85.56\% & 97.78\% \\
 & PSJS & 88.55\% & 88.28\% & 88.18\% & 76.77\% & 83.31\% & 80.63\% \\
 & AEM & 71.85\% & 73.33\% & 73.33\% & 53.70\% & 59.63\% & 58.89\% \\
 & Set F1 & 88.27\% & 86.82\% & 87.27\% & 73.32\% & 74.73\% & 76.02\% \\
\midrule
\multirow{4}{*}{politics} & EX & 91.79\% & 97.18\% & 98.21\% & 92.05\% & 97.69\% & 93.85\% \\
 & PSJS & 74.42\% & 81.66\% & 82.68\% & 53.76\% & 60.84\% & 71.92\% \\
 & AEM & 54.87\% & 63.33\% & 64.62\% & 40.00\% & 40.00\% & 53.33\% \\
 & Set F1 & 72.12\% & 79.49\% & 79.17\% & 54.11\% & 52.58\% & 70.04\% \\
\midrule
\multirow{4}{*}{\textbf{Average}} & EX & 95.44\% & 97.96\% & \textbf{98.68\%} & 91.01\% & 92.59\% & \textbf{95.79\%} \\
 & PSJS & 79.16\% & 85.36\% & \textbf{86.35\%} & 64.17\% & 71.96\% & \textbf{75.78\%} \\
 & AEM & 63.71\% & 69.25\% & \textbf{71.89\%} & 47.66\% & 54.17\% & \textbf{57.97\%} \\
 & Set F1 & 78.13\% & 81.18\% & \textbf{83.39\%} & 60.71\% & 66.45\% & \textbf{71.53\%} \\
\bottomrule
\end{tabular}
\end{table*}

\paragraph{Self-Correction Loop} The iterative correction loop improves the Cypher query by incorporating execution results, evaluator feedback, and optional named-entity validation. In the reported CypherBench configuration, the runner allows up to three query-generation attempts per question. At each step, the evaluator either accepts the query, routes it directly back to the generator with feedback, or invokes validation: labels, property values, and relationship patterns receive Levenshtein-similarity suggestions, while property values additionally receive LLM-ranked semantic suggestions before feedback aggregation. The refinement loop is detailed in Algorithm~\ref{alg:cypher_refinement}.

\paragraph{Implementation Details} \textit{VeriCypher} is implemented in Python~3.11+ using LangGraph for workflow orchestration and state management. The graph backend is Memgraph \cite{memgraph}, an in-memory, C++-backed, open-source graph database. LLM calls in the evaluated workflows (Table~\ref{table:workflow_metric_comparison}) use OpenRouter through LangGraph's client.

% ================================================================================================
\section{Experiments}\label{sec:experiments}
% ================================================================================================

\subsection{Experimental Setup}
% ------------------------------------------

\paragraph{Datasets}
% ------------------------------------------
We evaluate \textit{VeriCypher} on the test partition of CypherBench~\cite{DBLP:conf/acl/FengPR25}, a benchmark for question answering over Wikidata-derived property graphs using Cypher. The implemented sweep uses seven domains: \textit{company}, \textit{fictional character}, \textit{flight accident}, \textit{geography}, \textit{movie}, \textit{NBA}, and \textit{politics}. The corresponding JSON files contain 347, 385, 189, 366, 401, 270, and 390 question--answer pairs, respectively, for 2,348 examples in total.

Each CypherBench graph is converted to a Memgraph-compatible JSON representation and loaded before the corresponding run. The generator is conditioned on compact node and relationship schema summaries derived from the same graph. We therefore evaluate against the full available per-domain test files rather than a fixed-size random sample.

\begin{figure*}[!t]
\centering
\includegraphics[width=0.90\textwidth]{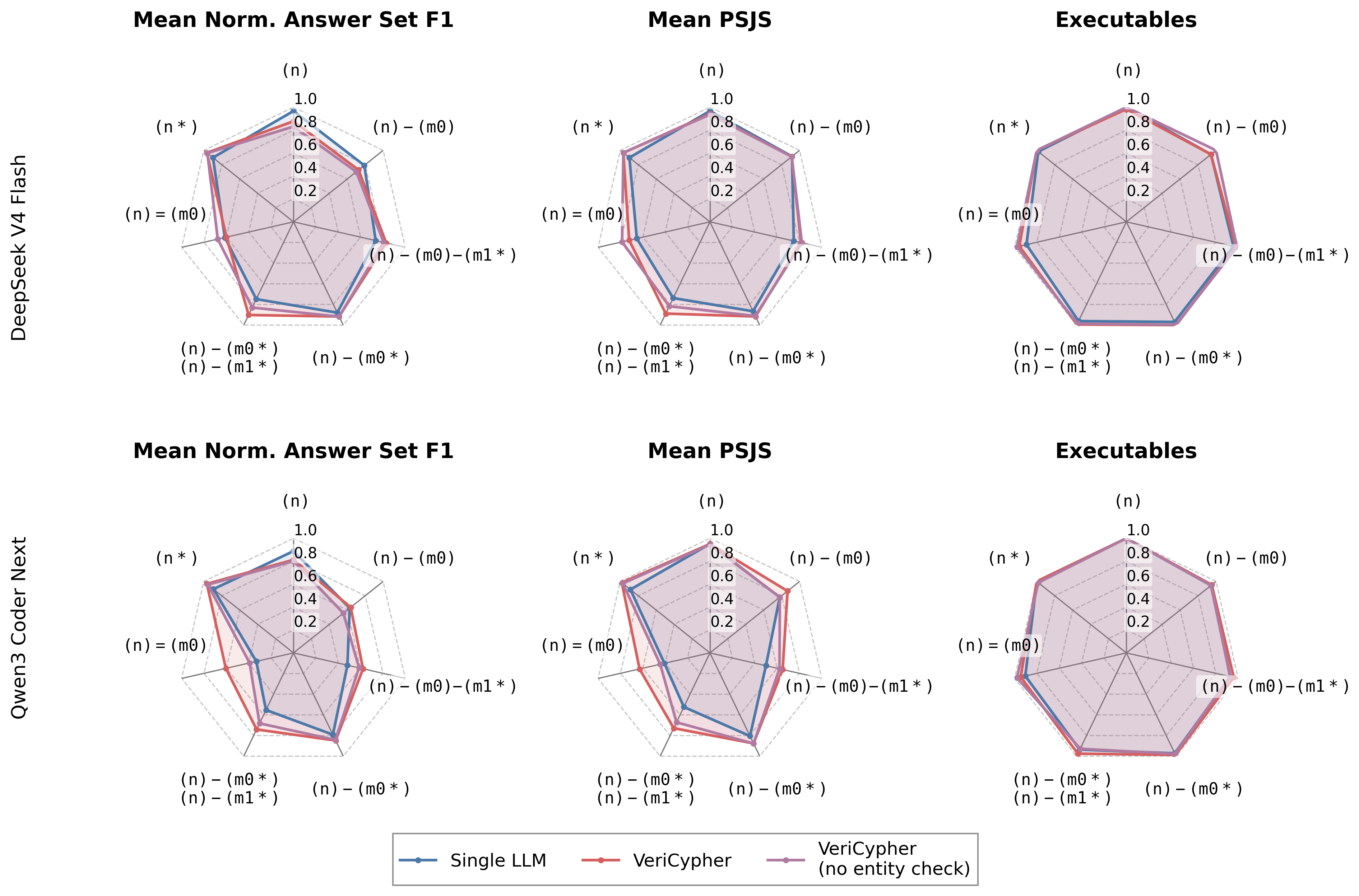}
\caption{\textbf{Basic MATCH patterns.} Core CypherBench metrics by basic MATCH-pattern template. \textit{Behavior:} \textit{VeriCypher} is most consistently helpful on relationship-heavy and repeated-node templates because these cases expose more opportunities for wrong relationship direction, missing edge constraints, or entity mismatch; execution feedback and entity validation can correct failures that a single-pass generator leaves unresolved.}
\label{fig:basic_match_spider}
\end{figure*}

Pattern-level results are analyzed in Fig.~\ref{fig:basic_match_spider}.

\paragraph{Baseline and Models}
% -------------------------------
We compare three implemented LangGraph workflows. \textbf{Single} is the \texttt{single\_llm} baseline and consists of Query Generator $\rightarrow$ Memgraph Executor. \textbf{VeriCypher} adds query evaluation, named-entity extraction and checking, fix-instruction generation, and feedback aggregation before regeneration. \textbf{No-entity-check} is an ablation in which evaluator feedback can trigger regeneration, but the validation subgraph is bypassed. In the current configuration, all LLM-backed nodes in a run use the same OpenRouter model: \textit{Qwen3 Coder Next} or \textit{DeepSeek V4 Flash}. We selected these two backbones because they represent strong code-oriented open-weight model families available through the same inference interface, allowing us to test whether the verification loop is robust across different LLM providers rather than tuned to a single model. The runner allows up to three query-generation attempts per question.

\paragraph{Evaluation Metrics}
% -------------------------------

We report the four CypherBench metrics used in the metric summary table. \textit{EX} measures executability: a generated Cypher query receives 1 if it runs successfully on the loaded Memgraph graph and 0 otherwise. \textit{PSJS} is provenance-subgraph Jaccard similarity, computed as the Jaccard overlap between the sets of graph nodes touched by the gold and predicted queries. \textit{AEM} is normalized answer exact match, a strict denotation metric that compares normalized predicted and gold answer tables. \textit{Set F1} is a relaxed denotation metric that computes precision, recall, and F1 over canonicalized answer values. All table entries are averaged over questions within each graph and reported as percentages.

\begin{figure*}[!t]
\centering
\includegraphics[width=0.90\textwidth]{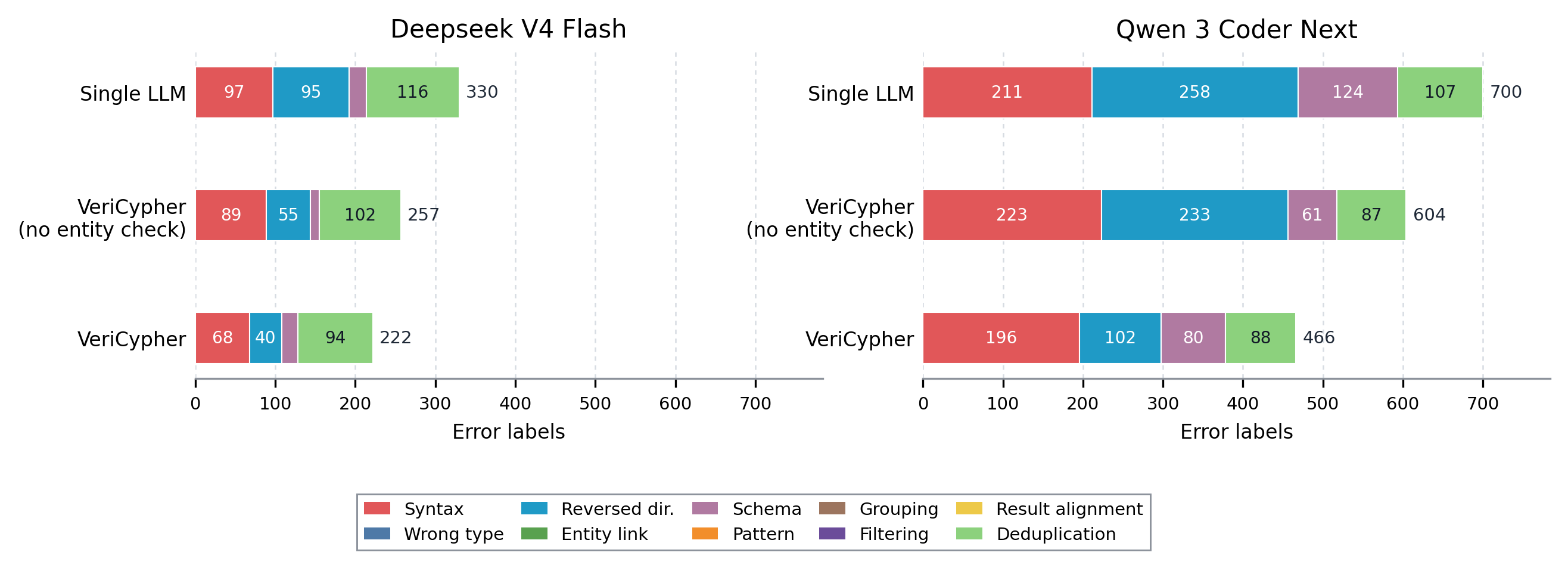}
\caption{\textbf{Error profile.} Aggregate error profile across CypherBench domains. Bar-end totals count assigned labels, not failed questions, because one incorrect example may receive multiple labels. \textit{Behavior:} remaining failures concentrate in semantic alignment and structurally complex cases because executability feedback is strong for syntax and schema errors, but disjunction, grouping, optional matching, and multi-intent questions require planning choices that local repair instructions do not always recover.}
\label{fig:error_analysis_aggregate}
\end{figure*}

\begin{figure*}[!t]
\centering
\includegraphics[width=0.90\textwidth]{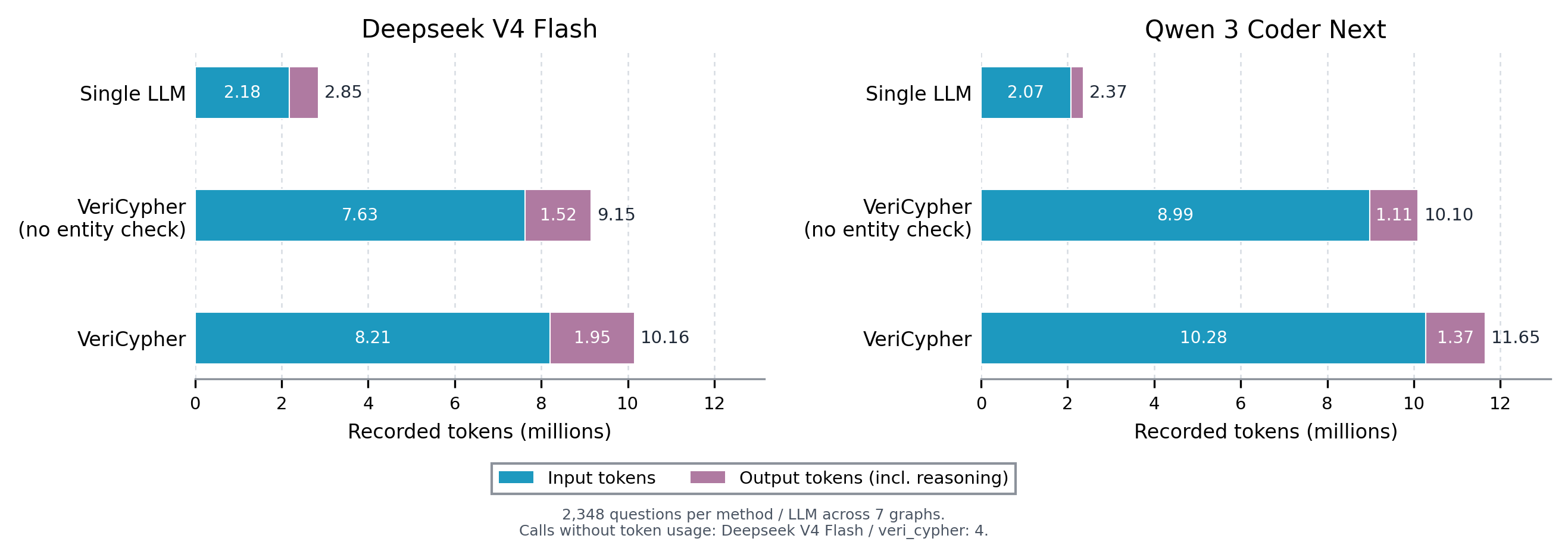}
\caption{\textbf{Token usage.} Aggregate recorded input and output tokens for three workflows and two LLM backbones, evaluated on the same 2,348 questions across seven CypherBench graphs. Full \textit{VeriCypher} uses more tokens than the single-pass baseline for both backbones. Historical tracking may omit failed invocations; four DeepSeek \textit{VeriCypher} calls lack token usage. These recorded totals measure inference usage rather than monetary cost.}
\label{fig:token_usage_aggregate}
\end{figure*}

\subsection{Experimental Results}
% ------------------------------------------
Table~\ref{table:workflow_metric_comparison} summarizes performance across seven CypherBench graphs, two LLM backbones, and three workflows. The full \textit{VeriCypher} workflow obtains the strongest average result for both backbones on all four reported metrics. For DeepSeek V4 Flash, it improves over the \textit{Single} baseline from 95.44\% to 98.68\% on EX, 79.16\% to 86.35\% on PSJS, 63.71\% to 71.89\% on AEM, and 78.13\% to 83.39\% on Set F1. For Qwen3 Coder Next, the corresponding gains are 91.01\% to 95.79\%, 64.17\% to 75.78\%, 47.66\% to 57.97\%, and 60.71\% to 71.53\%.

The ablation results show that execution and evaluator feedback alone are useful but not sufficient. The no-entity-check workflow often improves over \textit{Single}, especially on denotation metrics, yet it is less stable than the full system and can regress on particular graph--metric pairs. The full workflow recovers this gap by grounding candidate entities and relationship patterns against the database before regeneration, which is reflected in the consistently higher average PSJS and answer scores.

\subsection{Pattern and Error Analysis}
The analysis plots provide a more detailed view of this behavior. The basic MATCH-pattern breakdown in Fig.~\ref{fig:basic_match_spider} shows the largest and most consistent gains on relationship-heavy and repeated-node templates, where wrong edge direction, missing constraints, or entity mismatch are common. The error profile in Fig.~\ref{fig:error_analysis_aggregate} shows that the remaining errors concentrate in semantic alignment and structurally complex queries, indicating that the current feedback loop is effective for local schema and execution repair but less reliable when the question requires explicit planning over disjunctions, grouping, optional matching, or multiple intents.

\subsection{Token Usage}
The token-usage comparison is shown in Fig.~\ref{fig:token_usage_aggregate}. Across the same 2,348 questions, the full workflow uses more recorded tokens than the single-pass baseline for both LLM backbones. These token totals measure inference usage rather than monetary cost and may omit failed invocations.

\section{Discussion and Limitations}
% --------------------------------------
The experimental results suggest that \textit{VeriCypher}'s main benefit is not merely higher executability, but more faithful graph grounding. In Table~\ref{table:workflow_metric_comparison}, executability is already high for several single-pass runs, yet the full workflow still improves PSJS and answer-denotation metrics. This indicates that many baseline queries are syntactically valid but semantically misaligned with the intended graph evidence. The validation loop helps by turning execution and evaluator feedback into database-grounded repairs of labels, relationship patterns, property names, and literal entity values.

The ablation clarifies the role of the entity-checking subgraph. In the no-entity-check workflow, evaluator requests for entity validation are routed directly back to the query generator, without running named-entity extraction, database checks, or validation-feedback aggregation. Its gains over the single-pass baseline show that iterative execution feedback is useful on its own. However, the full workflow is more consistent across both LLM backbones, which supports the view that explicit database checks reduce hallucinated or mismatched graph elements rather than only prompting the model to try another query.

The pattern-level analysis in Fig.~\ref{fig:basic_match_spider} further shows where this mechanism matters most. Gains are strongest on relationship-heavy and repeated-node MATCH templates, where a query can run while still traversing the wrong relation, omitting a constraint, or binding the wrong entity. These cases benefit from a verifier that can expose concrete mismatches in the graph schema or database contents. Conversely, the aggregate error profile shows that the remaining failures are less dominated by simple executability errors and more by semantic alignment and structurally complex query formulation.

These results also reveal limitations. The current system performs local repair more reliably than global planning. When a question requires disjunction, grouping, optional matching, symmetric relationship handling, or multiple coordinated subgoals, a valid repair instruction may not be enough to recover the intended query structure within the attempt budget. Future work should therefore combine the present verification loop with explicit query planning, decomposition of multi-intent questions, and richer intermediate representations that separate entity grounding from higher-level Cypher program structure.

% =====================================================================================
\section{Conclusion}
% =====================================================================================
We presented \textit{VeriCypher}, an execution-grounded Text-to-Cypher framework for question answering over labeled property graphs. The framework combines query generation, execution, LLM-based evaluation, database-backed named-entity verification, fix-instruction generation, and feedback aggregation in a single refinement loop. On CypherBench, \textit{VeriCypher} improves both evaluated LLM backbones across executability, provenance overlap, and denotation metrics, indicating that verification helps generated Cypher queries become not only runnable but also better grounded in the intended graph evidence.

The ablation study shows that execution feedback is useful on its own, but explicit entity and schema grounding makes the refinement loop more reliable. The largest gains appear on relationship-heavy and repeated-node MATCH templates, where a query may execute successfully while still using the wrong relationship direction, omitting a constraint, or binding an incorrect entity. These results support named-entity verification as a practical mechanism for reducing graph-specific hallucinations in LLM-generated Cypher.

Remaining errors concentrate in semantic alignment and structurally complex queries. Future work will therefore focus on combining \textit{VeriCypher}'s verification loop with stronger query planning and subgoal decomposition, while extending evaluation to multi-turn settings and additional property-graph domains.

\section*{Acknowledgment}
OpenAI Codex was used for text review, code development assistance, and preparation of the LaTeX source.

% ==== References ====
\bibliographystyle{IEEEtran}
\bibliography{paper-main-bib}

\end{document}